%% file: ijcai23.tex
\documentclass{article}
\pdfoutput=1
\usepackage{ijcai23}

\usepackage{times}
\usepackage{soul}
\usepackage[normalem]{ulem}
\usepackage{url}
\usepackage[hidelinks]{hyperref}
\usepackage[utf8]{inputenc}
\usepackage[small]{caption}
\usepackage{graphicx}
\usepackage{amsmath}
\usepackage{amsthm}
\usepackage{booktabs}
\usepackage{algorithm}
\usepackage{algorithmic}
\usepackage[switch]{lineno}
\usepackage{verbatim}
\usepackage{color}

\title{XVTP3D: Cross-view Trajectory Prediction Using Shared 3D Queries \\for Autonomous Driving}

\author{
Zijian Song$^{1,4}$
\and
Huikun Bi$^2$\thanks{Corresponding Author. Code and models are released at \href{https://github.com/hasika000/xvtp3d}{https://github.com/hasika000/xvtp3d}}\and
Ruisi Zhang$^{3}$\and
Tianlu Mao$^1$\and
Zhaoqi Wang$^1$
\affiliations
$^1$Beijing Key Laboratory of Mobile Computing and Pervasive Device, \\Institute of Computing Technology, Chinese Academy of Sciences\\
$^2$SenseTime\\
$^3$Intuitive Surgical\\
$^4$University of Chinese Academy of Sciences
\emails
songzijian21s@ict.ac.cn,
bihuikun@senseauto.com,
Ruisi.Zhang@intusurg.com,
ltm@ict.ac.cn, 
zqwang@ict.ac.cn
}

\begin{document}

\maketitle

\begin{abstract}
Trajectory prediction with uncertainty is a critical and challenging task for autonomous driving. Nowadays, we can easily access sensor data represented in multiple views. 
However, cross-view consistency has not been evaluated by the existing models, which might lead to divergences between the multimodal predictions from different views. It is not practical and effective when the network does not comprehend the 3D scene, which could cause the downstream module in a dilemma. 
Instead, we predicts multimodal trajectories while maintaining cross-view consistency.
We presented a cross-view trajectory prediction method using shared 3D Queries (XVTP3D). We employ a set of 3D queries shared across views to generate multi-goals that are cross-view consistent. We also proposed a random mask method and coarse-to-fine cross-attention to capture robust cross-view features. As far as we know, this is the first work that introduces the outstanding top-down paradigm in BEV detection field to a trajectory prediction problem. The results of experiments on two publicly available datasets show that XVTP3D achieved state-of-the-art performance with consistent cross-view predictions.
\end{abstract}

\input{Sections/1-introduction/1-introduction.tex}

\input{Sections/2-relatedwork/2-relatedwork.tex}

\input{Sections/3-method/3-method.tex}
\input{Sections/4-experiments/4-experiments.tex}

\input{Sections/5-conclusion/5-conclusion.tex}

\bibliographystyle{named}
\bibliography{ijcai23}

\end{document}

%% file: Sections/1-introduction/1-introduction.tex
\section{Introduction}

\begin{figure}[t]
    \centering
    \includegraphics[scale=0.32]{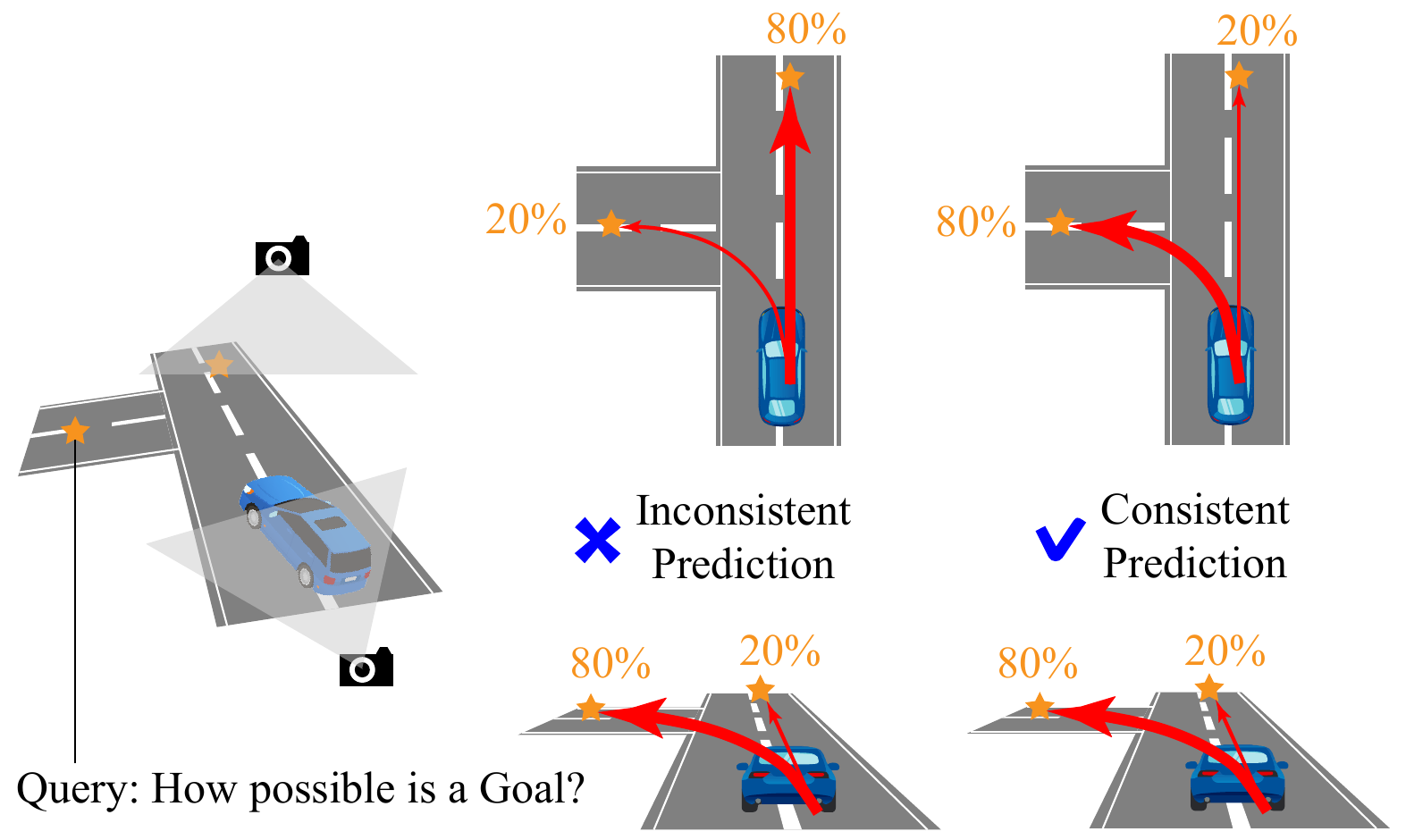}
    \caption{Driving environment can be observed via multiple views but the predicted goal in different views should be consistent. (Left) A traffic scene is observed in BEV and FPV. (Middle) The front candidate goal in BEV is assigned a higher probability than the left one, while the result is reversed in FPV. It is contradictory in reality. 
    (Right) A compliant prediction with consistently assigned probabilities between views.
    }
    \label{fig:introduction}
\end{figure}

Trajectory prediction for autonomous driving is challenging but critical. To improve the performance of autonomous driving, more sensors are used to collect comprehensive, real-time data for the driving environment. Point cloud data collected by LiDAR is used to capture the location of visible objects in 3D space, which can be projected onto a bird's-eye view representation. Camera-based data is represented in either the first-person or fixed-specific view. Based on the characteristics of each view, we can better perform current trajectory prediction to improve the performance of decision-making by analyzing the dynamic driving environment.


Existing trajectory prediction methods \cite{bi2019joint,lin2021vehicle,varadarajan2022multipath++,wang2022ltp} for autonomous driving have been able to achieve plausible results using single-view data. However, it is unclear how to effectively and efficiently utilize multi-view data. It is challenging to use the point cloud data from LiDAR, when it is projected onto a bird's-eye view, to track the continuous motion of objects. Even if it can be well proceeded, the performance cannot be guaranteed in 3D space, especially for the overlaying objects. For the first-person view, the spatial distance has errors in pixel precision when projected onto a 2D plane, and the faraway elements are easy to omit. Recent works have proposed trajectory prediction methods using multi-view data to overcome the limitation of only using single-view data. The aggregation of the cross-view features is primarily divided into two categories: feature extraction for each view separately followed by a concatenation of single-view features \cite{rasouli2019pie,yao2019egocentric,malla2020titan,bi2020can}, and adaptive extraction of temporal-spatial features in multi-view by attention-related modules \cite{yin2021multimodal}. The interaction and propagation of cross-view features used in these two methods are only preserved at the view or agent level, without considering the consistent correlations in 3D motion space. In the cross-view feature aggregation, the invalid information and errors from each view might be amplified and transferred to the other views, which leads to incorrect trajectory predictions (Fig.\,\ref{fig:introduction}).

We propose \textit{XVTP3D}, a cross-view trajectory prediction model using shared 3D queries. Unlike existing multi-view-based trajectory prediction work simply aggregate cross-view data,
we perform the trajectory prediction based on the characteristics of each agent's movement in 3D space, and the fusion of cross-view features:
(1) we develop a coarse-to-fine cross-view attention mechanism to extract features from the vectorized representation inputs targeting at agent and roadmap elements. A coarse attention layer is used to filter the interaction information with high weight for each single view. A fine attention layer is then used to enhance the fusion of the filtered interaction information across views. (2) We construct a group of 3D queries in motion space as predicted goal candidates, then map them to each view, respectively. We use this top-down paradigm to {generate} the unified 3D motion representation {from each 2D space in each single view}. This reduces the errors that come from predicting goals from a single view. (3) The multi-goals generated in a single view are mapped to the heatmap to ensure unified outputs
, and the cross-view multi-target that dominates in each view is chosen as the final motion trend. We also show that our XVTP3D has better performance for trajectory prediction from {each view}.

The main contributions of this work can be summarized as: 
{(1) We proposed XVTP3D using a coarse-to-fine cross-view attention mechanism with shared 3D queries. It is capable of capturing and aggregating cross-view characteristics to ensure the 3D consistent multi-goals.} 
(2) As our best knowledge, we are the first to introduce a top-down paradigm for cross-view features fusion in trajectory prediction. We validated our method in Argoverse dataset and Nuscenes dataset. The experimental results showed that our method can achieve state-of-the-art performance by generating multimodal predictions that are compliant, accurate, and rational.

%% file: Sections/2-relatedwork/2-relatedwork.tex
\section{Related Work}
This section reviews trajectory prediction work related to our work. 

\input{Sections/2-relatedwork/2-1MotionForecasting}
\input{Sections/2-relatedwork/2-2CrossViewPrediction}
\input{Sections/2-relatedwork/2-33DObjectDetection}

%% file: Sections/2-relatedwork/2-1MotionForecasting.tex
\subsection{Trajectory Prediction}
The future motion of an agent is constantly influenced by the occurrences in the scene, including surrounding agents, and the external environment. For trajectory prediction, it is essential to model complex social interactions.
Early work proposed using GNN to model interaction.
STGAT \cite{huang2019stgat} first used a spatial-temporal graph to capture the interaction between agents.
Following work attempted to use various of GNN such as graph transformer network \cite{yu2020spatio}, graph convolutional neural network \cite{mohamed2020social}, and message passing graph neural network \cite{choi2021shared}.
Other work, such as SGCN \cite{shi2021sgcn} built a sparse directed graph based on their reckon that the dense undirected graph would introduce superfluous interactions. To account for environmental impact, recent work \cite{gao2020vectornet,zhao2020tnt,gu2021densetnt,liang2020learning,kim2021lapred,wang2022ltp,zhou2022hivt} used vectorized representations due to their superior performance over rasterized representations\cite{chai2019multipath,gilles2021home,gilles2022gohome,messaoud2021trajectory}.

Multimodality, which refers to the possibility of different future motions based on the same historical observation, is an additional crucial challenge in trajectory prediction.
There are two primary approaches: implicit and explicit.
The implicit methods provided for the potential intent of the target agent by utilizing latent variables. They generally employ generative networks to model the joint data distribution such as using GAN \cite{gupta2018social,dendorfer2021mg}, VAE \cite{lee2017desire,salzmann2020trajectron++,chen2021personalized,choi2021shared,chen2022scept}, and diffusion \cite{gu2022stochastic}.
The explicit methods selected multiple potential future motion from a set of anchors \cite{phan2020covernet,zhao2020tnt,gu2021densetnt,varadarajan2022multipath++,wang2022ltp,sun2022m2i}.
Each anchor explicitly represents a potential future motion.

Both explicit and implicit methods are incapable of generating multiple meaningful modes.
The mode collapse problem adversely affects implicit techniques (also called the posterior collapse problem in VAE). Regardless of the context, the majority of their anticipated outputs are limited to a small number of comparable modes, typically one or two.
In addition, their results cannot be interpreted as randomly generated. Some efforts have tried to solve this problem \cite{shao2020controlvae,choi2021shared}, but yet no major progress.
The explicit methods can provide highly interpretable results, but their multimodal predictions are limited. They are limited to only one or two important modes. In our work,  we intend to develop a more relevant future mode by using cross-view information.

%% file: Sections/2-relatedwork/2-2CrossViewPrediction.tex
\subsection{Cross-view Trajectory Prediction}
The majority of trajectory prediction research is conducted based on a single view. It is difficult to generalize these frameworks to fit the cross-view scenario. There are a few cross-view methods.
The first cross-view trajectory prediction datasets was introduced by Tsotsos's research group.
They collected and released the PIE and JAAD datasets \cite{rasouli2019pie,rasouli2017they}, which contain the trajectories of both BEV and FPV images using a ego-vehicle.
They proposed a prediction method across the two views.
Their work was extended by TITAN \cite{malla2020titan} and LOKI \cite{girase2021loki}, which further introduced the external environment and radar data to improve the prediction accuracy.
To better integrate information from the two views, Yao et al. \cite{yao2019egocentric} proposed a pooling mechanism based on ROI, Bi et al. \cite{bi2020can} stacked two streams of RNN to encode observations from the two views independently and then concatenated them to obtain a joint social feature. Yin et al. \cite{yin2021multimodal} employed a transformer network to combine the dynamic feature across two views.

These methods considered observations from multiple views resulting in significant improvements, which showed the importance of cross-view information.
However, the existing cross-view prediction methods have shortcomings:
(1) the potentially inconsistent between views due to lack of consideration of the consistency, 
(2) lack of capability of multi-modal modeling, since the functions is designed to superimpose 2D features and cannot effectively represent the 3D scene (i.e., they usually apply specific aggregation functions to aggregate features from multiple views, such as pooling (yao 2019 egocentric), concatenation \cite{bi2020can}, and attention mechanism \cite{yin2021multimodal}),
(3) high computational cost due to the required additional prior knowledge \cite{rasouli2019pie,casas2018intentnet,malla2020titan}.

%% file: Sections/2-relatedwork/2-33DObjectDetection.tex
\subsection{3D Object Detection}
The goal of 3D object detection is to predict the 3D bounding boxes of each object in the scene based on images observed from multiple views.
The traditional methods detected objects in a bottom-up manner.
They extract features in each view (i.e., bottom) then fuse their result in 3D space (i.e., up).
Experiments showed that this traditional paradigm is sensitive to compound error \cite{wang2022detr3d}.
Recent work proposed novel top-down paradigms for 3D object detection. DETR3D \cite{wang2022detr3d} used a set of 3D queries in 3D space (i.e., top); each 3D query represented a predicted bounding box reference point.
These 3D queries were projected to each view (i.e., down) to interact with their feature map.
The refined 3D queries constituted an effective representation of the 3D scene.
PETR \cite{liu2022petr} eliminated the projection and the interpolation in DETR3D to reduce computational complexity.
BEVFormer \cite{li2022bevformer} used a temporal dimension to maintain temporal stability.
FUTR3D \cite{chen2022futr3d} expanded DETR3D's applicability to accommodate a greater variety of sensor data.
The top-down paradigm works much better than the bottom-up paradigm, and achieves remarkable success in 3D object detection. We use the top-down paradigm for our cross-view trajectory prediction for autonomous driving because it is similar to how 3D information is captured.

%% file: Sections/3-method/3-method.tex
\section{Methodology}

\begin{figure*}[htbp]
    \centering
    \includegraphics[scale=0.38]{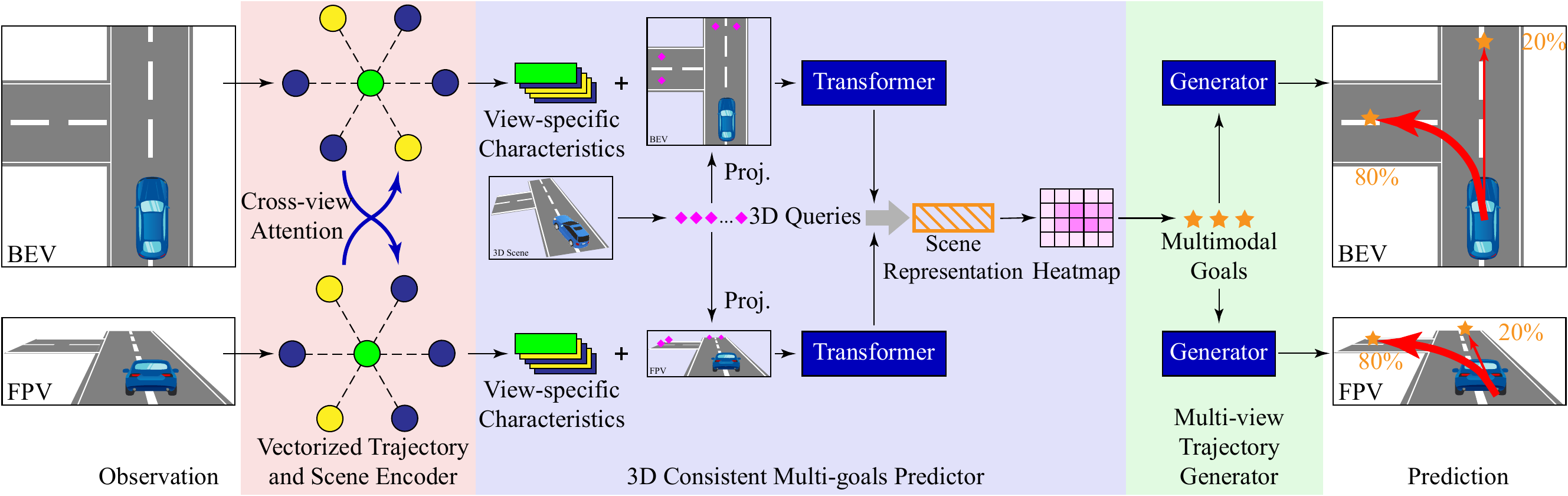}
    \caption{Overall framework of XVTP3D. Our architecture is a goal-driven method. It is composed of three parts: (a) Vectorized Trajectory and Scene Encoder includes a BEV branch and a FPV branch. (b) 3D Consistent Multi-Goals Predictor generates consistent cross-view goals prediction using shared 3D queries. It first transforms the 3D queries into a scene representation using the encoded features. It then predicts a heatmap and samples multimodal goals. (c) Multi-view Trajectory Generator generates the trajectories in each view {accordingly with the predicted multi-goals.} 
    }
    \label{fig:2}
\end{figure*}

\input{Sections/3-method/3-1ProblemFormulation}
\input{Sections/3-method/3-2OverallFramework}

\input{Sections/3-method/3-3SceneEncoder}
\input{Sections/3-method/3-4GoalPredictor}
\input{Sections/3-method/3-5TrajectoryCompleter}
\input{Sections/3-method/3-6ImplementationDetails}

%% file: Sections/3-method/3-1ProblemFormulation.tex
\subsection{Problem Formulation}

We generalize the trajectory prediction task as follows. A scenario comprises $N$ agents and a static environment. The historical state sequence $S$ comprises each agent's past positions and velocities. The static environment comprises lane semantics ($E$).
The objective is to predict the future trajectory $\hat{Y_{i}}$, where the ground truth is $Y_{i}$, for the target agent $i \in[1,N]$ at the time period from $t+1$ to $(t+T_{pred})$, given the historical observation $ X= \left(E, S\right) $ from $(t-T_{obs}+1)$ to $t$.



We propose a cross-view trajectory prediction method to perform a cross-view motion transformation by characterizing temporal and spatial relationships between distinct perspectives. For a given traffic scenario from $M$ different views, meaning $M$ observations are accumulated as $ \left\{ X_{m_1}, X_{m_2}, ..., X_{m_M} \right\} $. Our approach is to simultaneously predict the future trajectory of each target agent $i$ under the corresponding view $ \left\{ \hat{Y}_{i, m_1}, \hat{Y}_{i, m_2}, ..., \hat{Y}_{i, m_M} \right\} $.



We use $m$ to identify each view observation.
In this paper, we consider two views: the bird's-eye view, and the first-person view, where $M=\left\{bev,fpv\right\}$. We noted our proposed framework can be applied to a larger range of situations, not limited to BEV and/or FPV.

%% file: Sections/3-method/3-2OverallFramework.tex
\subsection{Overall Framework}

We propose a cross-view trajectory prediction method (XVTP3D) using shared 3D Queries. Our XVTP3D model (Fig.\,\ref{fig:2}) comprises (1) a scene encoder (Sec.\,3.3) utilizing vectorized inputs, including two encoder branches for BEV/FPV, respectively, to characterize each perspective, (2) 3D consistent multi-goals predictor (Sec.\,3.4) utilizing the encoded features and a set of cross-view goals to characterize the uncertainty of future motion, while preserving the consistency between views based on the 3D queries that are shared across views, (3) multi-view trajectory generator (Sec.\,3.5) utilizing the predicted cross-view goal for each agent to produce predicted trajectories in each view.

%% file: Sections/3-method/3-3SceneEncoder.tex
\subsection{Vectorized Trajectory and Scene Encoder}

Our method utilizes two identically structured encoders for each view, named BEV Encoder, and FPV Encoder. They are designed to encode vectorized observation within their respective views using VectorNet \cite{gao2020vectornet} (which is a hierarchical graph neural network for subgraph and global-graph). We develop a coarse-to-fine cross-view attention mechanism to capture social interactions, which preserved the consistency and relatively significance within each view.

\subsubsection{Vectorized Representation}
We use subgraph to extract dependence from each instance's vectorized representation. It refines the input vectors within each instance (e.g. agent, lane) and outputs their attributes $a_{i}$ through an output layer.
\begin{equation}
    v^{\left( l+1 \right)}_{i}=MLP\left(\left[ v_{i}^{\left( l \right)}; \varphi _{attn}\left( v_{i}^{\left( l \right)}, v_{i}^{\left( l \right)}, v_{i}^{\left( l \right)} \right) \right]\right)
\end{equation}
\begin{equation}
    a_{i} = \varphi _{agg}\left(v^{\left(L\right)}_{i}\right)
\end{equation}
where $MLP$ denotes a multi-layer perceptron, $l$ denotes the $l$-th layer of the total $L$ refinement layers, $\left[;\right]$ denotes concatenation, $\varphi _{agg}$ denotes max-pooling, and $\varphi _{attn}$ denotes multi-head attention. 

We use global-graph to capture the interaction across the entire scene. The global interaction is represented by a fully connected graph. Each instance in the graph corresponds to a vertex. Their attributes $a_{i}$ are assigned to the vertices. The global-graph refines the vertex attribute 
between instances for the $l$-th layer:
\begin{equation}
    a_{i}^{\left( l+1 \right)}=MLP\left(\left[ a_{i}^{\left( l \right)}; \varphi _{attn}\left( a_{i}^{\left( l \right)}, a_{\mathcal{N}}^{\left( l \right)}, a_{\mathcal{N}}^{\left( l \right)} \right) \right]\right)
\end{equation}
\begin{equation}
    p_{i} = a_{i}^{(L)}
\end{equation}
where $\mathcal{N}$ denotes instances in the scene other than $i$,
$\varphi _{attn}$ denotes multi-head attention.
The attributes at the last (i.e., $L$-th) layer are used as the final state features.

We compute the dot-product attention for each head within multi-head attention for both subgraph and global-graph. The multi-head attention outputs the concatenation of all heads {whose QKV are interpreted as vertex attributes}:
\begin{equation}
    Attention\left( Q,K,V \right) =soft\max \left( \frac{QK^T}{\sqrt{d_k}}V \right) 
\end{equation}

\subsubsection{Coarse-to-fine Cross-view Attention}
The attention mechanism extracts the semantic information of social interactions. The attention weight $\alpha_{ij}$ indicates the relatively significance of instance $j$ to instance $i$:
\begin{equation}
 \alpha _{ij}=\frac{\exp \left( a_i\cdot a_{j}^{T} \right)}{\sum_{k \in \mathcal{N}}{\exp \left( a_i\cdot a_{k}^{T} \right)}}
\end{equation}

To further extract the semantic information across views, we propose a coarse-to-fine cross-view attention mechanism (Fig.\,\ref{fig:2}). We {modified}
the original attention mechanism in global-graph using the cross-view attention module. It enhances the important instances to ensure that each potentially influential instance is considered. Concretely, we employ a coarse attention layer to identify significant instance cross-views and a fine attention layer to calculate the final result on the previously chosen instance set.

The coarse attention layer determines the initial attention weights by calculating the attention weights in each view ($\left\{\alpha_{ij, bev}|j \in \mathcal{N}\right\}$ and $\left\{\alpha_{ij, fpv}|j \in \mathcal{N}\right\}$). Based on attention weights and predefined thresholds ($\varepsilon$), we focus on certain instances in each view and proceed further matching and filtering on them.
\vspace{-0.0cm}
\begin{equation}
 \mathcal{N}^*_{bev}=\left\{ j|j\in N, \alpha_{ij, bev} >\varepsilon \right\}
\end{equation}
\begin{equation}
 \mathcal{N}^*_{fpv}=\left\{ j|j\in N, \alpha_{ij, fpv} >\varepsilon \right\}
\end{equation}

The observed data from each view contains characteristics related to occlusion, perspective, or distance, and the learned attention weights derived from both views, which are not necessarily comparable. Considering the consistency of cross-view motion, we take the instances focused in either view as the final instance set: 

\begin{equation}
 \mathcal{N}^{*} = \mathcal{N}^*_{bev} \cup \mathcal{N}^*_{fpv}
\label{eq:instance_set}
\end{equation}
A fine layer refines vertex attribute on the instance set (Eq.\,\ref{eq:instance_set}):
\begin{equation}
 \varphi _{attn}\left( a_i,a_{\mathcal{N}},a_{\mathcal{N}} \right) =\sum_{k \in \mathcal{N}^{*}}{\frac{\exp \left( a_i\cdot a_{j}^{T} \right)}{\sum_{k\in \mathcal{N}^{*}}{\exp \left( a_i\cdot a_{k}^{T} \right)}}} \cdot a_{j}
\end{equation}

\begin{figure}
    \centering
    \includegraphics[scale=0.75]{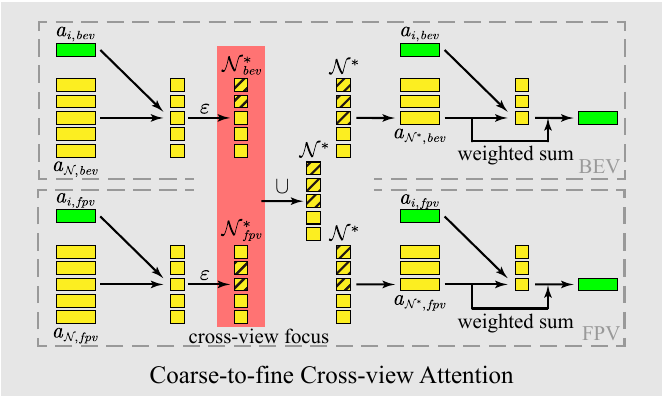}
    \caption{Coarse-to-fine Cross-view Attention module. It computes the Multi-Head Attention of the coarse/fine layer to incorporate social information from BEV and FPV (denoted by dashed boxes). Leveraging the "union operation" (denoted by the red box), it takes into account the consistency of interaction between two views resulting in a more precise allocation of attention weights.}
    \label{fig:cv_attention}
\end{figure}

\subsubsection{Encoder Loss}
We utilize the sparse goal scoring loss \cite{gu2021densetnt} to accelerate the convergence of our encoder.
We employ an MLP to score the sparse goal, and compute the classification loss:
\begin{equation}
 \sigma _{sparse}=soft\max \left( MLP\left( p ^{\left(L\right)} \right) \right) 
\end{equation}
\begin{equation}
 \mathcal{L}_1=CE\left( \sigma _{sparse} , \pi _{sparse} \right) 
\end{equation}
where $CE$ denotes the cross entropy, and
$\pi _{sparse}$ denotes the ground truth sparse goal scoring (one for ground truth goal sparse, and zero for others).

%% file: Sections/3-method/3-4GoalPredictor.tex
\subsection{3D Consistent Multi-goals Predictor}
We propose a multi-goals predictor to generate the cross-view goals while preserving the consistency between views using a set of 3D queries, which are employed as goal candidates (Fig.\,\ref{fig:2}). Those are transformed into an effective scene representation. Inspired by DETR3D \cite{wang2022detr3d} developed for other cross-view tasks \cite{wang2022detr3d,liu2022petr,li2022bevformer,chen2022futr3d}, we use a similar top-down paradigm. Unlike traditional bottom-up approaches directly concatenated view-specific characteristics, we transform them into a 3D motion space via shared 3D queries. Then, we score each 3D query, and project them onto a 2D heatmap. Because our 3D queries are shared across views, the predicted heatmap and motion are guaranteed to be consistent across views. Last, we sample the multimodal goals based on the predicted heatmap. 

\begin{figure}
    \centering
    \includegraphics[scale=0.75]{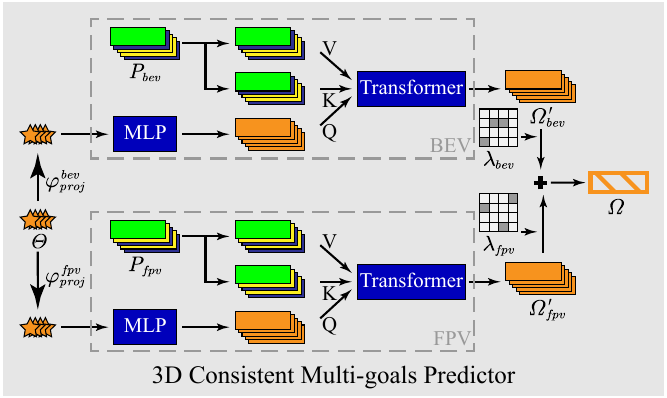}
    \caption{The 3D Consistent Multi-goals Predictor. The 3D Queries (orange stars) carry view-specific features (green boxes) to 3D space, forming a unified scene representation (the box with orange diagonal). 3D Queries are first projected onto two views, then calculated by Transformer in each view, avoiding the error interference and accumulation between the views.}
    \label{fig:goal_predictor}
\end{figure}

\subsubsection{Generating and Projecting 3D Queries}
Inspired by DETR3D \cite{wang2022detr3d}, we initiate a set of 3D queries in 3D space $\varTheta = \left\{ \theta_{1}, \theta_{2}, \dots, \theta_{n} \right\}$.
Each 3D query represents a goal candidate in the 3D scene $\theta \in \mathbf{R}^{3}$. A query can be basically interpreted as asking how likely this point is to become a goal. It is sampled using an anchor-free algorithm \cite{gu2021densetnt}, which introduces sparse and dense goals. We registered them in a 3D space and our queries can shared across views to forecast consistent cross-view motions.

In the transformation procedure $\mathcal{F}$, we compute an effective representation for each query using the encoded features:
\begin{equation}
    \varOmega=\left\{ \omega_{1}, \omega_{2}, ..., \omega_{n} \right\}=\mathcal{F}\left(\varTheta, P_{bev}, P_{fpv}\right)
\end{equation}
We use the top-down paradigm (1) to pass the 3D queries (projected to each view) through a transformer along with view-specific features, (2) to pass the computed features to a 3D motion space using 3D queries and merge them into a unified representation: 
\begin{equation}
    \varOmega _{bev}^{\prime}=TF\left( MLP\left( \varphi _{proj}^{bev}\left( \varTheta \right) \right) ,P_{bev},P_{bev} \right) 
\end{equation}
\begin{equation}
    \varOmega _{fpv}^{\prime}=TF\left( MLP\left( \varphi _{proj}^{fpv}\left( \varTheta \right) \right) ,P_{fpv},P_{fpv} \right) 
\end{equation}
\begin{equation}
   \varOmega =\varOmega _{bev}^{\prime}\odot \lambda _{bev}+\varOmega _{fpv}^{\prime}\odot \lambda _{fpv} 
\end{equation}
where $P_{m}=\left\{p_{1, m}, p_{2, m}, ..., p_{i, m}\right\}$ denotes all state features in view $m$, $\varphi _{proj}$ denote the projection transformation from camera's intrinsics and extrinsics to the corresponding view, $\odot$ denotes the element-wise product, and $\lambda$ denote mask matrix for the corresponding views. It masks the invisible query (e.g., the candidate out of sight) to avoid the effect of invisible query to the final outputs. 
$TF$ denotes a Transformer layer that comprises a multi-head attention, a feed forward neural network, and a layer of normalization with residual connection successively. Transformer's computing elements $\left\{Q,K,V\right\}$ is equivalent of multi-head attention's $\left\{Q,K,V\right\}$.
   \begin{equation}
    TF\left( Q, K, V \right) =LN\left( FFN\left( Attention\left( Q, K, V \right) \right) \right) 
\end{equation}

\subsubsection{Heatmap Scoring}
The network $\mathcal{F}$ transforms the original 3D queries into an effective representation of the 3D scene with cross-view information. After that, we employ an MLP as an output layer to score each query. The classification loss between the predicted scores and ground truth scores is computed as:
\begin{equation}
    \sigma _{goal}=soft\max \left( MLP\left( \varOmega \right) \right) 
\end{equation}
\begin{equation}
    \mathcal{L}_2=CE\left( \sigma _{goal} , \pi _{goal} \right) 
\end{equation}
where $CE$ denotes the cross entropy, and $\pi_{goal}$
denotes the score of the ground truth (one for reaching the endpoint of a given trajectory, and zero for all others).

The goal distribution is presented by the heatmap to show all the candidates and their scores. Using the hill climbing algorithm \cite{gu2021densetnt}, we sample a set of goals $G=\left\{g_{1}, g_{2}, ..., g_{k}\right\}$ from the heatmap. Our multimodal goals $G$ are plausible and consistent across views because the 3D query and heatmap are shared.

During training, the network may favor one view over the other causing it to be degenerated into a single-view network. To avoid this and enhance cross-view robustness, we generate a random mask as $\omega$. The original mask matrix was generated merely based on the visibility of a given candidate. We use the random probability $\beta = 0.1$ to mask some visible candidates. We perform this random mask independently on each view, meaning the masked candidates are asymmetrical. Some candidates can be concealed in one view but apparent in the other. The asymmetric data improves the robustness.

%% file: Sections/3-method/3-5TrajectoryCompleter.tex
\begin{table*}[htbp]
\centering
\resizebox{0.7\textwidth}{!}{
\begin{tabular}{cccccccc}
\toprule
\multicolumn{4}{c}{Method} & \multicolumn{2}{c}{Argoverse} & \multicolumn{2}{c}{Nuscenes} \\ 
\makebox[0.04\textwidth][c]{No.} & \makebox[0.04\textwidth][c]{Que} & \makebox[0.04\textwidth][c]{RM} & \makebox[0.04\textwidth][c]{CA} & \makebox[0.12\textwidth][c]{BEV} & \makebox[0.12\textwidth][c]{FPV} & \makebox[0.12\textwidth][c]{BEV} & \makebox[0.12\textwidth][c]{FPV} \\
\midrule
1 & - & - & -        & 0.980/1.138   & 21.39/24.27   & 2.483/4.753   & 26.94/50.95  \\
2 & + & - & -        & 0.876/1.123   & 18.09/18.03   & 2.299/4.295   & 24.21/45.01  \\
3 & - & - & +        & 0.790/1.101   & 20.12/23.23   & 2.360/4.632   & 25.49/47.82  \\
4 & + & + & -        & 0.740/1.032   & 17.53/\textbf{15.60}   & 2.003/3.539   & 22.19/37.73  \\
5 & + & + & +        & \textbf{0.733/1.030}   & \textbf{16.06}/16.25   & \textbf{1.907/3.115}   & \textbf{19.99/32.65}  \\
\bottomrule 
\end{tabular}
}
\caption{Results of our ablation studies. We report minADE/minFDE in BEV and FPV. Lower is better.}

\label{table:ablationstudy}
\end{table*}

\begin{table*}[htbp]
\centering
\resizebox{0.5\textwidth}{!}{
\begin{tabular}{ccccccc}
\toprule
$\beta$ & 0.0         & 0.5         & 1.0         & 1.5         & 2.0         & 2.5         \\
\midrule 
BEV  & 0.77/1.09   & 0.78/1.06   & 0.73/1.03   & 0.74,1.05  & 0.85/1.08   & 0.80/1.10   \\
FPV  & 16.0/17.6 & 16.9/16.9 & 16.1,16.3 & 16.6,16.6 & 18.7/17.1 & 17.7,18.4 \\
\bottomrule 
\end{tabular}
}
\caption{Influence of random probability $\beta$ on Argoverse dataset (minADE/minFDE).}

\label{table:randomprobability}
\end{table*}

\subsection{Multi-view Trajectory Generator}

We use a multi-view trajectory generator to predict multimodal trajectories in each view. Each trajectory is conditional on each predicted goal in the 3D consistent multi-goals predictor. Similar to our encoder, our trajectory generator contains two decoding branches, named BEV generator and FPV generator, with identical structure but independent parameters.

We project the predicted goals onto each respective view. To capture the high-dimensional features of each goal, we use an MLP as an embedding layer. Embedding features, and state feature of each target agent $i$ are concatenated as joint features, which are passed through an MLP to generate the complete trajectories: 
\begin{equation}
    \hat{Y}_{i, bev}=MLP\left( \left[ p_{i, bev}; MLP\left( \varphi _{proj}^{bev}\left( g_{i, k} \right) \right) \right] \right) 
\end{equation}
\begin{equation}
    \hat{Y}_{i, fpv}=MLP\left( \left[ p_{i, fpv}; MLP\left( \varphi _{proj}^{fpv}\left( g_{i, k} \right) \right) \right] \right) 
\end{equation}
where $\varphi _{proj}$ denotes the projection transformation to the corresponding view, $k$ denotes the $k$-th multimodal prediction generated by the $k$-th goal.

\subsubsection{Regression Loss}
We compute the Euler distance (i.e, L2 loss) between the predicted trajectory and the ground truth trajectory, using also the teacher forcing technique \cite{gao2020vectornet}:
   \begin{equation}
    \mathcal{L}_3=\sum_i{\left\| \hat{Y}_i-Y_i \right\| _2}.
\end{equation}

%% file: Sections/3-method/3-6ImplementationDetails.tex
\subsection{Implementation Details}


\paragraph{Losses.}
The loss function of each view is a weighted sum of three loss terms. We use the sum of the losses under each view as our objective function:
\begin{equation}
    \mathcal{L}=\sum_{m\in M}{w_{m}\mathcal{L}_{m}}=\sum_{m\in M}{w_{m}\left(w_1\mathcal{L}_1+w_2\mathcal{L}_2+w_3\mathcal{L}_3\right)}
\end{equation}
where $\mathcal{L}_1$, $\mathcal{L}_2$, and $\mathcal{L}_3$, respectively, represent sparse goal scoring loss of the encoder, cross-view goal scoring loss, and trajectory regression loss.
$w_1, w_2, w_3$ are weights to balance three terms, and
$w_{m}, m \in M$ are weights to balance the multiple views.

In our implementation, the subgraph and the global-graph have six refinement layers $L=6$. All the mentioned multi-head attentions in our network comprise four attention heads. The embedding size is 128. All the MLPs/Transformers in our framework share a similar structure but independent parameters, which contain two layers with a hidden size of 256. The probability of a random mask is $\beta = 0.1$ as it perform best.

%% file: Sections/4-experiments/4-experiments.tex
\renewcommand{\dblfloatpagefraction}{0.9}
\section{Experiment}

We compare our XVTP3D model with state-of-the-art prediction methods for autonomous driving and present quantitative and qualitative evaluation results in this section.

\begin{table*}[htbp]
\centering

\resizebox{0.65\textwidth}{!}{
\begin{tabular}{ccccc}
\toprule
 & \multicolumn{2}{c}{Argovers} & \multicolumn{2}{c}{Nuscenes} \\
\makebox[0.14\textwidth][c]{Method} & \makebox[0.12\textwidth][c]{BEV} & \makebox[0.12\textwidth][c]{FPV} &
\makebox[0.12\textwidth][c]{BEV} & \makebox[0.12\textwidth][c]{FPV} \\
\midrule 
Traj++        & 0.90/1.92 & 33.70/52.19 & 1.88/3.65 & 25.83/57.28 \\
DenseTNT      & 0.73/1.05 & 18.21/\underline{23.36} & 2.52/4.80 & 28.11/50.48 \\
LaPred        & 0.71/1.44 & \underline{17.88}/26.14 & \textbf{1.53}/\underline{3.37} & \underline{21.43}/\underline{42.72} \\
HiVT          & \textbf{0.69}/\underline{1.04} & $\--$ / $\--$ & $\--$ / $\--$ & $\--$ / $\--$ \\
Ours          & \underline{0.73}/\textbf{1.03} & \textbf{16.06/16.25} & \underline{1.91}/\textbf{3.11} & \textbf{19.99/32.65} \\
\bottomrule 
\end{tabular}
}
\caption{Quantitative comparison on the Argoverse/Nuscenes dataset. We report the minADE/minFDE in BEV and FPV. Bold and underlined numbers indicate the best and the second best, respectively.}

\label{table:comparison}
\end{table*}

\input{Sections/4-experiments/4-1ExperimentalSetup}

\input{Sections/4-experiments/4-2AblationStudies}
\input{Sections/4-experiments/4-3QuantitativeEvaluation}
\input{Sections/4-experiments/4-4QualitativeEvaluation}

%% file: Sections/4-experiments/4-1ExperimentalSetup.tex

\paragraph{Dataset.}
We use two publicly accessible datasets for autonomous driving: Argoverse \cite{chang2019argoverse} and NuScenes \cite{caesar2020nuscenes}. Both datasets provide agent trajectories and vectorized scene geometry.
They also provide other auxiliary information, including semantic labels, camera intrinsics and extrinsics. 
The Argoverse dataset \cite{chang2019argoverse} contains 323557 samples and is split into 205942 for training, 39472 for validation, and 78143 for testing.
The length of all the sequences is 5 seconds. We use the first 2 seconds for observations, and the last 3 seconds for predictions.
The sample rate of the Argoverse dataset is 10 Hz. The Nuscenes dataset \cite{caesar2020nuscenes} contains 32186 samples for training and 9041 samples for validation.
The total length of the sequence is 8 seconds. We use the first 2 seconds for observations, and the last 6 seconds for predictions.
The sample rate of the Nuscenes dataset is 2 Hz.

\paragraph{Data Processing for More Views.}
Both the Argoverse and Nuscenes datasets provide bird's-eye-view motion data and scene geometry.
To get the observations from other views, we project the original instances onto the first-person view of the target agent by using the known intrinsics and extrinsics of the front camera.
To eliminate the effect of ego-future, we also use the absolute coordinate introduced by \cite{choi2021shared}. For each sequence, the coordinate system is fixed at the starting position of the target agent, facing forward, which did not respond to the agent's ego-motion.
In FPV, we retain the same semantic labels as BEV.
Training samples with long and unseen future trajectories are detrimental to the training process.
For this reason, we eliminated the training data that contained unqualified future lengths, which is approximately 3$\%$ of the total amount.

\paragraph{Metrics.}
We choose \textbf{minADE} and \textbf{minFDE} as our evaluation metrics, which are the two major metrics adopted in the Argoverse competition and the Nuscenes competition.
\textbf{ADE} (Average Displacement Error) is the average L2 distance between the ground truth and our prediction over all predicted time steps.
\textbf{FDE} (Final Displacement Error) is the distance between the predicted final destination and the true final destination at the end of the prediction period.
Both lowering the ADE and FDE are preferable.
The prefix \textbf{min} stands for a multimodal evaluation strategy. That is, we sample $k=6$ possible trajectories and calculate metrics based on the trajectory that is closest to the ground truth. Note that the following experiments focuses more on the minFDE results than its minADE results. Our method is built on a goal-driven framework. It is convenient to describe the consistency of 3D cross-view motion by emphasizing the trajectory's endpoints as opposed to the whole trajectory.

\paragraph{Baselines.}
We compared our method to four state-of-the-art trajectory prediction models for autonomous driving (i.e., baselines): Traj++ \cite{salzmann2020trajectron++}, DenseTNT \cite{gu2021densetnt}, LaPred \cite{kim2021lapred}, HiVT \cite{zhou2022hivt}. The results of Traj++, DenseTNT, LaPred, and HiVT are generated from the released code. We evaluated all methods on both BEV and FPV. Due to the lack of data processing in the released code for the NuScenes dataset, the prediction of the first-person view requires separate processing of the target agents, while HiVT predicts all agents in the whole scene simultaneously, making separation challenging. Therefore, for HiVT, we only show the results of BEV for the Argoverse dataset. 





%% file: Sections/4-experiments/4-2AblationStudies.tex
\begin{figure*}[htbp]
    \centering
\includegraphics[scale=0.5]{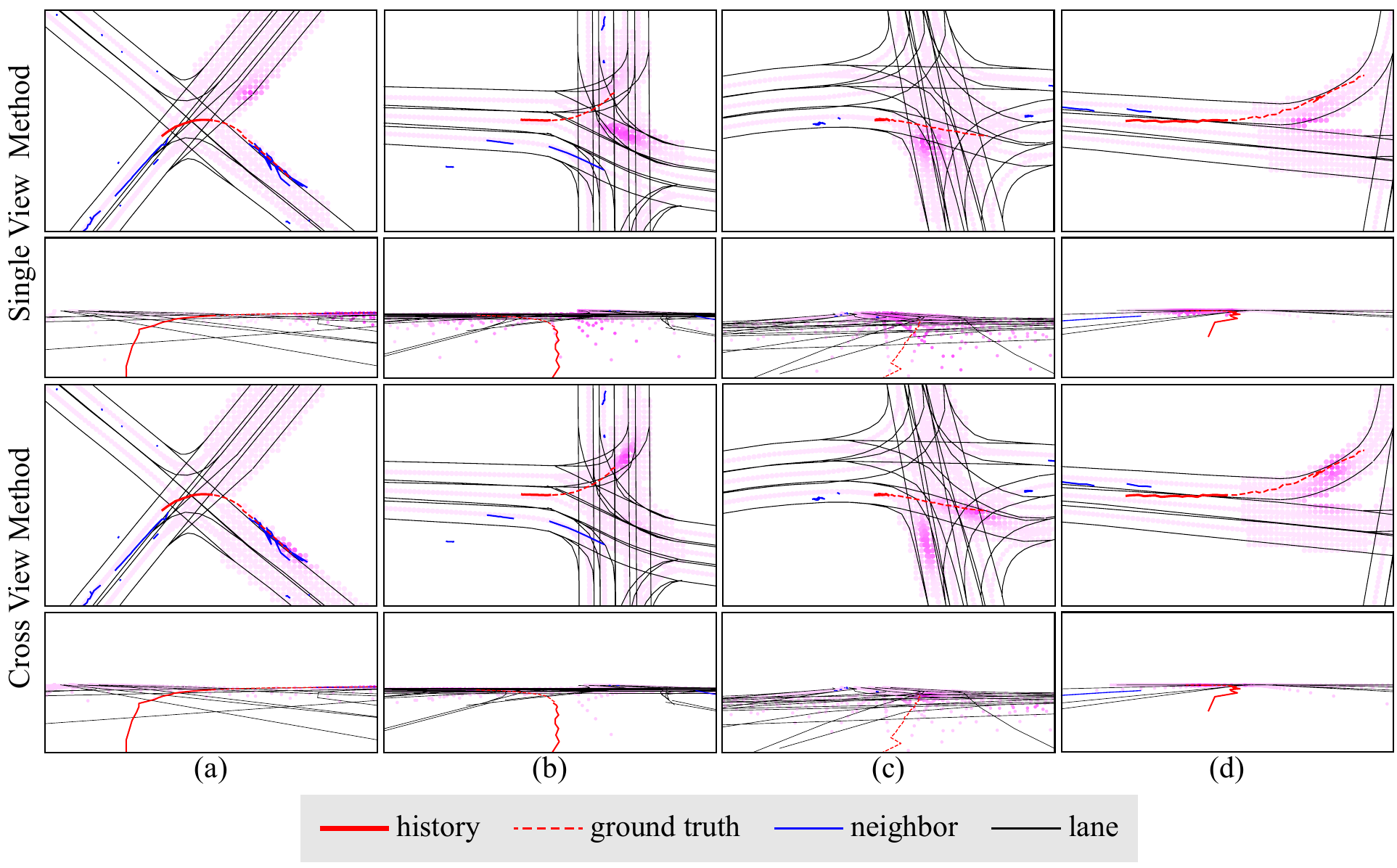}
    \caption{Illustration of cross-view consistency. Distribution is represented by pink dots. Darker color indicates higher probability, and vice versa. Each column shows a case from (a) to (d). The first two rows show the inconsistent heatmap predicted by the single view method in BEV and FPV. The last two rows show the consistent heatmap predicted by our method in BEV and FPV. The top rows do not have consistency constraints, while the bottom rows have consistency constraints. In the top of (a), the results in BEV and FPV are inconsistent, with the BEV outputting a dark area in the front indicating a prediction of going straight and the FPV outputting dark areas turning right. This suggests that single-view methods are susceptible to producing unreasonable predictions due to noise interference. At the bottom of (a), our method ensures consistent constrain to rectify the inaccurate prediction in BEV by the accurate prediction in FPV, resulting in a more reasonable prediction without errors. (b) shows a scenario similar to (a). In the top of (c), the BEV outputs a dark area in the right, indicating a prediction of turning right, while the FPV outputs a dark area in the front, indicating a prediction of going straight. BEV and FPV are inconsistent and have a single-mode problem. This indicates that single-view methods are susceptible to mode collapsing. In the bottom of (c), our method accepts both two plausible possibility provided by BEV and FPV, allowing us to produce a more diverse prediction. (d) shows a scenario similar to (c).} 
    \label{fig:heatmap}
\end{figure*}

\subsection{Ablation Study}
We conduct ablation studies on both the Argoverse dataset and the Nuscenes dataset.
We isolate the contribution of each module by gradually removing it.
We use the symbols ``+'' and ``-'' to represent whether we add or drop these components.
We especially investigate the following three components: (1) \textbf{Que}: Que denotes the 3D consistent multi-goals predictor using the shared 3D queries. ``-'' indicates predicting goals on each single-view, we refer to it as the single-view method. (2) \textbf{RM}: denotes the random mask technique employed during mask matrix generation. (3) \textbf{CA}: denotes the cross-view attention in our encoder.

The quantitative results of our ablation studies are presented in Table.\,\ref{table:ablationstudy}. And we can conclude the contribution of each component as follows:




(1) \emph{Cross-view Predictor.} The 3D consistent multi-goals predictor takes view-specific characteristics of each view into the 3D space and merges them into a unified representation of the 3D scene. This is entirely distinct from the single-view approach, which provides prediction in two views independently. The first and second rows of results indicate the efficacy of our cross-view predictor. The single-view method (Method\,1) has the highest error in prediction. The prediction accuracy increased by $9.6 \% \sim 10.6 \%$ in BEV and $11.7 \% \sim 28.7 \%$ in FPV when multiple views are combined in Method\,2. The improvement in FPV is greater due to the fact that the FPV provides limited information for prediction. The improvement of the two views supports our motivation to utilize cross-view data.
The single-view approaches capture just single-view trajectories; it does not contain cross-view information and is therefore incapable of comprehending 3D scenes.
In contrast, our cross-view predictor can leverage the cross-view information more, which effectively generates a reasonable 3D scene representation.

(2) \emph{Cross-view Consistency.} The 3D consistent multi-goals predictor connects two views, while maintaining 
cross-view consistency by creating multimodal goals that are shared between views. The comparison between Methods\,2 and 3 demonstrates the significance of cross-view consistency. The approach with the cross-view attention module (Method\,3) can also connect and retrieve cross-view features from two views. It outperforms the single-view method (Method\,1) in BEV by $2.5\%\sim3.3\%$ in BEV and $6.1\%\sim8.1\%$ in FPV. However, it still generates multimodal predictions that are nonconforming and lack cross-view consistency.
This explains why the CA approach (Method\,3) in the Nuscenes dataset performs $7.3\%$ worse than the Que method (Method 2) in BEV and $5.9\%$ worse in FPV. The aforementioned results suggested that, with the benefit of cross-view consistency, our method might more accurately predict trends, particularly in long-term predictions such as NuScenes.

(3) \emph{Random Mask.}  Table.\,\ref{table:ablationstudy} showed the random mask approach (Method 4) has the most substantial effect on performance. Adding the random mask increased the prediction accuracy by $8.1\%\sim17.6\%$ in BEV and $13.5\%\sim 16.2\%$ in FPV compared with Method\,2. Randomizing the mask can improve both views' performance. This shows that the original method employing a fixed mask matrix degenerates into an excessive reliance on one view, resulting in a weak performance for both views. By randomly masking instances, the predictor is prompted to extract more efficient cross-view features from multiple views, resulting in improved performance. Furthermore, we use a varying $\beta$, gradually increasing from 0.05 to 0.2 to find the best random probability $\beta$. Based on the performance shown in Table.\,\ref{table:randomprobability}, we use $\beta = 0.1$ in our final model.

(4) \emph{Cross-attention.} 
Increasing the cross-attention contribution to the final model improved the accuracy by $2.5\%\sim3.3\%$ in BEV and $6.1\%\sim8.1\%$ in FPV.  
As shown in the fourth and fifth rows of the Nuscenes dataset, dropping the cross-attention module reduced the final performance by $12.0 \%$ in BEV and $13.5 \%$ in FPV.  This showed that the cross-attention module can benefit our model by focusing more on significant instances and then encoding useful characteristics from different views. The cross-attention module did not add significantly to the overall performance of the Argoverse dataset. We believe this is because Nuscenes has a longer prediction length than Argoverse (3 seconds vs. 6 seconds).

%% file: Sections/4-experiments/4-3QuantitativeEvaluation.tex
\subsection{Quantitative Evaluation}

The quantitative results compared with baselines are reported in Table.\,\ref{table:comparison}. Traj++ performs poorly in Argoverse as it relies on rasterized HD maps, whereas the Argoverse dataset provides maps of driveable areas.

Table.\,\ref{table:comparison} showed our XVTP3D method outperforms other state-of-the-art methods in both BEV and FPV.
In terms of minFDE in BEV, our method achieves state-of-the-art performance for the Argoverse dataset against the other methods including HiVT, LaPred, DenseTNT.
For Nuscenes dataset, our method also achieves competitive accuracy compared with LaPred, with an improvement of $7.7 \%$. Our method significantly outperforms other methods in terms of minFDE.
Compared to LaPred, our method improves minFDE by $37.8 \%$ for the Argoverse and $23.6 \%$ for Nuscenes datasets.
Compared to DenseTNT, our method improves minFDE by $30.4 \%$ for the Argoverse and $35.3 \%$ for Nuscenes datasets. It suggested our method might be better at comprehending 3D scenes and generating more accurate predictions as a result. Our method achieved the highest precision in FPV while keeping outperformed precision in BEV.

%% file: Sections/4-experiments/4-4QualitativeEvaluation.tex
\begin{figure*}[htbp]
    \centering
\includegraphics[scale=0.5]{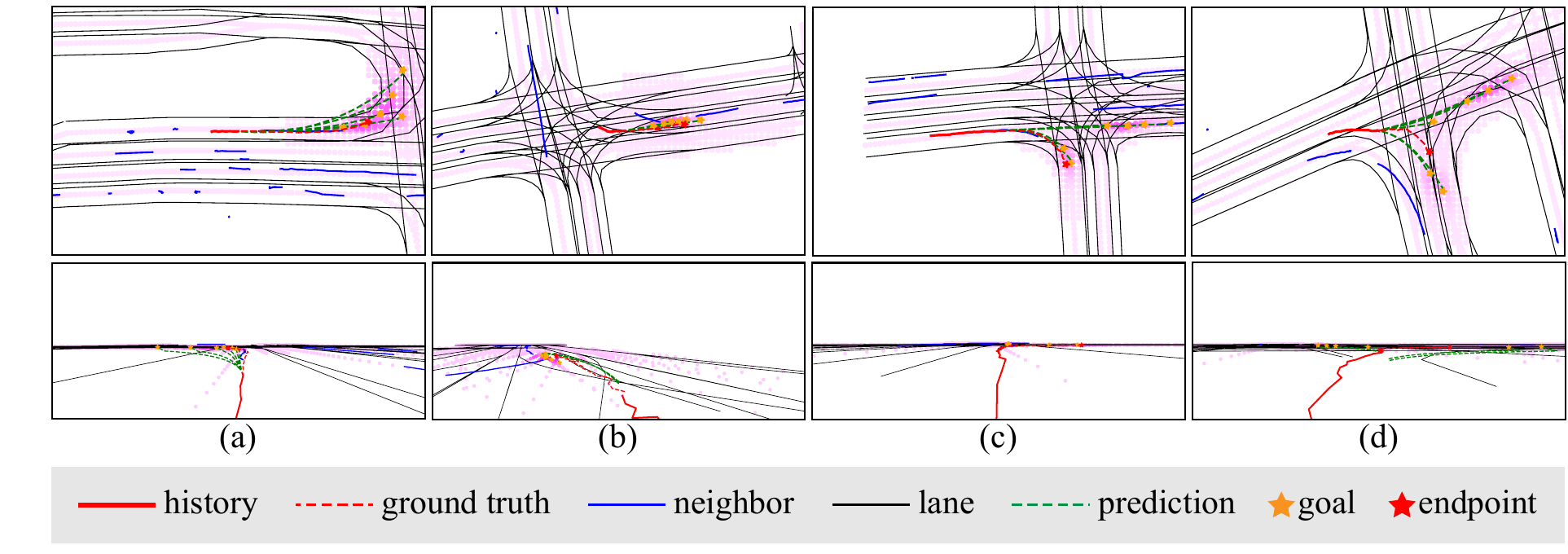}
    \caption{The visualization on Argoverse dataset. We presented our predicted goals as well as our predicted trajectories in four complex traffic scenarios. The first row shows the prediction in BEV, while the second row shows the prediction in FPV in the same scenario. We also present a failure case in the last column.
    }
    
\label{fig:trajectoryvisualize}
\end{figure*}

\subsection{Qualitative Evaluation}

\subsubsection{Heatmap Visualization}
Fig.\,\ref{fig:heatmap} qualitatively showed how cross-view consistency improves performance. We showed the predicted heatmap from both the single-view method and our full model in four scenes. Each column shows the prediction in the same scene for comparison. The top two rows show the BEV/FPV heatmaps generated by the single-view method, and the bottom two rows show the BEV/FPV heatmaps generated by our XVTP3D method. Pink dots represent the goal distribution. Deep color indicates high probability.

By imposing constraints on the consistency of the crossed views, our XVTP3D approach is able to combine the two views and adaptively choose the correct one.

In the first scenario (a), the BEV and FPV forecasts of the single-view method mismatch. In BEV, it is predicted that the target agent would go straight because the frontal region has a high likelihood.
In FPV, however, the target agent is more likely to turn right because of the higher probability assigned to the right region. Our model accepts the FPV results and rejects the BEV results. This aligns the two views so that the ground-truth trajectory can be used to make accurate predictions. The second row (b) shows a similar situation.
The agent is anticipated to make a right turn in BEV.
And in FPV,  it is predicted to turn left or right.
Our method corrects the prediction of BEV according to the result of FPV.

With the benefit of cross-view consistency, our method can generate more plausible future motion. In the third case (c), the results of the two views are also incompatible. The BEV predicts that the target agent will turn right while the FPV predicts that the target agent will go straight. Our method accepts both results to generate more multimodal predictions.
The right-turn mode is not visible in the FPV as it is outside the field of view. Despite the fact that this potential future motion is undetectable, our method continues to account for it. This substantiates our random mask strategy. In a similar way, the fourth row (d) shows an example of how our method takes two results from two views and makes a more accurate multimodal prediction.

\subsubsection{Trajectory Visualization}
We visualize our predicted trajectories on the Argoverse dataset in Fig.\,\ref{fig:trajectoryvisualize}. 
We illustrate various traffic scenarios, including left-turn (a), straight forward (b), and multiple possible future (c and d).
They are all difficult situations due to their location at a complex crossroads with several neighbors.
In the majority of complex driving scenarios (a and c) and from all views, our model generates accurate, multimodal, and reasonable predictions.
In the final column, we also present a failure case where our method predicts two plausible future motions and misses a deceleration.

%% file: Sections/5-conclusion/5-conclusion.tex
\section{Conclusion}
We proposed XVTP3D, a goal-driven approach to preserving cross-view consistency. Our method can predict trajectory using all accessible data from different views, which is not limited to BEV and FPV. The experimental results showed our proposed method achieved state-of-the-art performance.

\section*{Acknowledgements}
This work was supported in part by the National Natural Science Foundation of China under Grant 62002345, in part by the Innovation Research Program of ICT CAS (E261070).